\documentclass{article}

\usepackage[preprint,nonatbib]{neurips_2024}
\usepackage{graphicx, amsmath, amssymb, algorithm, algpseudocode, booktabs, lipsum}
\usepackage{wrapfig}

\usepackage[utf8]{inputenc} 
\usepackage[T1]{fontenc}    
\usepackage{hyperref}       
\usepackage{url}            
\usepackage{booktabs}       
\usepackage{amsfonts}       
\usepackage{nicefrac}       
\usepackage{microtype}      
\usepackage{xcolor}         
\usepackage{tabularx}

\title{FashionR2R: Texture-preserving Rendered-to-Real Image Translation with Diffusion Models}

\author{%
  Rui Hu\textsuperscript{1}\thanks{Work done during an internship at Style3D Research.} \,\thanks{These authors contributed equally to this work.}\ \ , \ 
  Qian He\textsuperscript{2}\footnotemark[2]\ \ , \ 
  Gaofeng He\textsuperscript{2}\,, \ 
  Jiedong Zhuang\textsuperscript{1}\,,\\
  \textbf{
  Huang Chen\textsuperscript{2}\,, \ 
  Huafeng Liu\textsuperscript{1}\thanks{Corresponding author.}\ \ , \ 
  Huamin Wang\textsuperscript{2}
  }
  \\
  \textsuperscript{1}Zhejiang University, \qquad
  \textsuperscript{2}Style3D Research \\
  \tt \small \textbf{
        \href{https://rickhh.github.io/FashionR2R/}{https://rickhh.github.io/FashionR2R/}
    }
}

\begin{document}

\maketitle

\begin{abstract}
Modeling and producing lifelike clothed human images has attracted researchers' attention from different areas for decades, with the complexity from highly articulated and structured content. Rendering algorithms decompose and simulate the imaging process of a camera, while are limited by the accuracy of modeled variables and the efficiency of computation. Generative models can produce impressively vivid human images, however still lacking in controllability and editability. This paper studies photorealism enhancement of rendered images, leveraging generative power from diffusion models on the controlled basis of rendering. We introduce a novel framework to translate rendered images into their realistic counterparts, which consists of two stages: Domain Knowledge Injection (DKI) and Realistic Image Generation (RIG). In DKI, we adopt positive (real) domain finetuning and negative (rendered) domain embedding to inject knowledge into a pretrained Text-to-image (T2I) diffusion model. In RIG, we generate the realistic image corresponding to the input rendered image, with a Texture-preserving Attention Control (TAC) to preserve fine-grained clothing textures, exploiting the decoupled features encoded in the UNet structure. Additionally, we introduce SynFashion dataset, featuring high-quality digital clothing images with diverse textures. Extensive experimental results demonstrate the superiority and effectiveness of our method in rendered-to-real image translation. 

\end{abstract}

\section{Introduction}
Modeling and simulating digital humans and clothing has achieved significant progress~\cite{wood2021fake, srivastava2024wordrobe, wu2022gpu, wang2023stable, li2023diffavatar}, while leveraging these 3D assets for fashion e-commerce still remains a challenging problem. Due to the imperfection of 3D models and the approximation in rendering algorithms, rendered images cannot yet replace fashion photos taken by a camera, with deficiency in the realism of rendered human faces and skin, clothing shape and fabric, etc. This paper studies transferring rendered fashion images into their realistic counterparts, which is inherently an Image-to-Image (I2I) translation problem.

Existing works on improving the realism of rendered images mainly resort to retrieving and blending real image patches~\cite{johnson2010cg2real}, or train a GAN-based network~\cite{bi2019deep, richter2022enhancing, wang2023enhancing} due to lack of paired training data. 
Another line of works can tackle this problem as general I2I translation~\cite{park2020contrastive, xie2023unpaired, kim2023unpaired, cheng2023general}. 
However, these methods may still suffer from several limitations: Firstly, their image generation pipelines have limited power to utilize real image resources for highly-detailed enhancement and may suffer from instability and mode collapse from adversarial training. Moreover, they either focus on indoor/outdoor scene enhancement while keeping coarse object-level semantic layout, or try to maintain face identity in training through loss constraints on sketches, and thus have difficulty in preserving fine-grained texture in clothing images. 

In this paper, we propose a novel framework based on diffusion models for rendered-to-real fashion image translation to address above limitations. Our main idea consists of two aspects: Firstly, we propose to leverage abundant generative prior from pretrained Text-to-Image (T2I) diffusion models~\cite{rombach2022high}, and apply simple adaptation to realistic image generation under the guidance of distilled rendered prior. Secondly, we adopt a texture-preserving mechanism by extracting spatial image structure through attention from an inversion pipeline.

To achieve this, we design a diffusion-based method consisting of two stages: Domain Knowledge Injection (DKI) and Realistic Image Generation (RIG). During DKI, we first finetune a pretrained T2I diffusion model~\cite{rombach2022high} on real fashion photos with derived captions from BLIP~\cite{li2022blip}, to adapt its capability in generating high-quality images to our target domain. After this adaptation, we propose to guide the image generation towards the negative direction of rendered effect. Inspired by Textual Inversion~\cite{gal2022image}, we distill a general rendered "concept" with thousands of rendered fashion images by training a negative domain embedding vector based on the adapted base model. 
During RIG, we employ a DDIM inversion~\cite{song2020denoising} pipeline to first invert a rendered image into the latent noise map, and then generate its corresponding real image using the previous embedding as a negative guidance~\cite{ho2022classifier}. Similar to recent training-free controls in T2I generation method~\cite{tumanyan2023plug, hertz2022prompt, cao2023masactrl, liu2024towards}, we discover that the attention map in the shallow layers of the UNet contains rich spatial image structure and can be used for fine-grained texture-preserving during the generation. Specifically, we inject query and key of the self-attention from the rendered image inversion and generation pipeline to the rendered-to-real image generation pipeline. This largely improves the consistency of intricate clothing texture details.

We evaluate our method on a public rendered Face Synthetics dataset~\cite{wood2021fake} and our collected SynFashion Dataset with fine-grained digital clothing and abundant texture variations. Empirical results comparing to previous works and experimental analysis demonstrate the efficacy of our method. Our main contributions are three-folds:
\\ (1) We propose a novel framework to address rendered-to-real fashion image translation by utilizing generative prior from pretrained diffusion models.
\\ (2) We inject rendered-to-real domain knowledge into a pretrained T2I diffusion model through positive domain finetuning and negative domain embedding, and design a texture-preserving attention control to preserve fine-grained clothing textures during the translation.
\\ (3) We collect a high-quality rendered fashion image dataset using the professional design software Style3D Studio, and plan to release the data with our paper to promote research in this important area.

\section{Related Works}

\subsection{Rendered-to-real Image Translation}
Improving the realism of rendered images has been a long-standing problem due to the inherent limitations of rendering pipelines and the rich potential for commercial applications. 
CG2Real~\cite{johnson2010cg2real} proposes to retrieve similar images from a large collection of real photos and then applies local style transfer to upgrade color, tone and texture of the CG image. Deep CG2Real~\cite{bi2019deep} adopts a two-stage deep learning framework to first transfer OpenGL images to PBR (Physically-Based Rendering) images, and then translates PBR to real images, disentangling lighting and texture in a CycleGAN-like~\cite{zhu2017unpaired} framework.~\cite{richter2022enhancing} enhances photorealism under the guidance of a set of input G-buffers and learns the network with a perceptual discriminator.~\cite{wang2023enhancing} proposes to learn a rendered image generator for human faces, which can encode the same face identity but different "style" from a real face image generator, based on StyleGAN~\cite{karras2020analyzing, karras2019style}. 
These methods all utilize limited data for generative training, while we propose to adapt diffusion models pretrained on large datasets for better image generation quality. Besides, applying these methods to fashion images often leads to the failure to preserve fine-grained clothing textures.

\subsection{Image-to-image Translation}
Transferring a rendered fashion image into its realistic counterpart is inherently an image-to-image (I2I) translation problem, which has attracted wide interest in different realms of research~\cite{chen2009sketch2photo, raad2017efros, shih2013data, wang2018high, richardson2021encoding}. Pix2Pix~\cite{isola2017image} utilizes a conditional-GAN~\cite{goodfellow2014generative, mirza2014conditional} and applies pixel-wise regularization based on paired training data, which is unavailable in many problem settings. Cycle-GAN~\cite{zhu2017unpaired} proposes to utilize cycle consistency~\cite{liu2017unsupervised, kim2017learning, yi2017dualgan, xu2024cyclenet} and optimizes a two-sided mapping between input source domain and output target domain. CUT~\cite{park2020contrastive} addresses the computational redundancy and over-restriction in this framework by simplifying it to one-sided~\cite{fu2019geometry, benaim2017one, xie2023unpaired} and introduces a patch-wise contrastive loss~\cite{jung2022exploring, wang2021instance, zheng2021spatially} for refined local constraints. UNSB~\cite{kim2023unpaired} proposes an iterative refinement method based on Schr\"odinger bridge to overcome potential mode collapse in GAN generation, while still has difficulty in faithfully translating high-resolution images. Different from general I2I tasks and domain adaptation, our method focuses on photorealism enhancement and can utilize more target-domain real photos for high-quality generation training, and thus can deal with imbalanced source-target training set. Style transfer~\cite{wang2004efficient, gatys2016image, gatys2017controlling} is a specific type of I2I task and can manage to transfer input source image to an arbitrary style~\cite{cheng2023general, li2017universal, huang2017arbitrary, deng2020arbitrary, zhang2023inversion} given one/few-shot target domain images as reference. These methods mainly focus on transferring style attributes like semantics, brushstrokes, colors, or material, while rendered-to-real requires preserving and enhancing complicated fine-grained details. Human/portrait relighting~\cite{Lagunas2021humanrelighting, yeh2022learning} modifies the nuanced lighting condition in the input image, while does not focus on enhancing realism and should leave geometry and materials untouched. Super-resolution methods~\cite{wu2023seesr, yu2024scaling, wang2023exploiting, chen2024low, saharia2022image,  liang2021swinir} address detail enhancement, while their success largely relies on synthesizing pseudo low-resolution images to obtain training pairs~\cite{wang2021real, zhang2021designing}, which is non-trivial for rendered-to-real problem.

\subsection{Diffusion-based Image Synthesis}
Recent progress in Text-to-Image (T2I) generation~\cite{rombach2022high, ramesh2022hierarchical, saharia2022photorealistic} based on diffusion models~\cite{sohl2015deep, ho2020denoising, dhariwal2021diffusion} opens up new opportunity for advancing rendered-to-real image translation. Many works have explored the possibility of utilizing abundant generative prior in pretrained diffusion models. 
Some~\cite{ruiz2023dreambooth, gal2022image} apply the adaptation of generation for a new concept with a few images, through either finetuning the base model~\cite{ruiz2023dreambooth}, or optimizing a text embedding~\cite{gal2022image}. Others~\cite{kawar2023imagic, brooks2023instructpix2pix} leverage text as guidance to edit a given image. However, rendered-to-real translation lies in the nuance of changes, which is too subtle to define as a "concept" or to capture with a few images. 
\cite{youwang2024paint, casas2023smplitex} leverage diffusion models for texture estimation or PBR synthesis, while mainly focusing on the generation of certain variables for the rendering pipeline, rather than subtle modification of preset variables in a given input image. 
Additionally,~\cite{tumanyan2023plug, hertz2022prompt, cao2023masactrl, liu2024towards} discover that the attention in the SD UNet captures rich image features and can apply to content preservation and modification. In our work, we utilize self-attention in shallow layers from the rendered image inversion, to impose the consistency of fine-grained texture in image translation.

\begin{figure}        
\centering
\includegraphics[width=\textwidth]{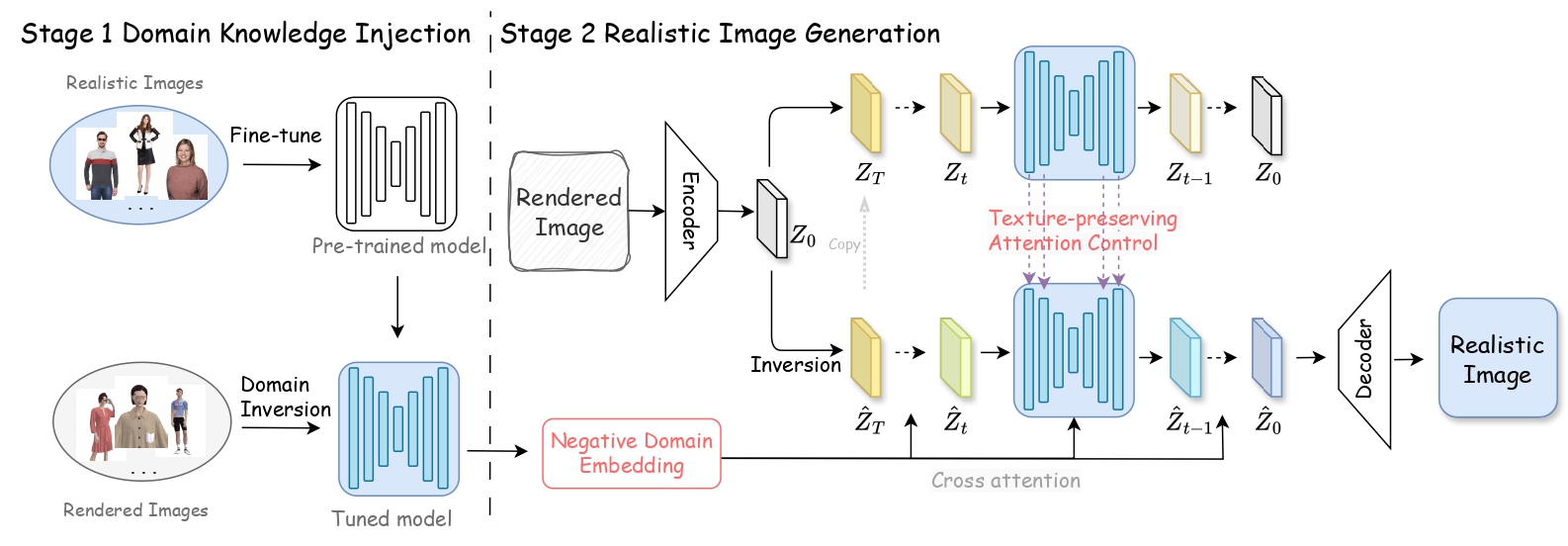} 
\caption{The overall pipeline of our proposed method.}
\label{method}                     
\end{figure}

\section{Method}
\subsection{Preliminaries}
\subsubsubsection{\textbf{Latent Diffusion Models.}}
In diffusion framework, the forward diffusion process begins by generating noisy images $x_t$ from clean images $x_0$ sampled from a specified data distribution, accompanied by their respective noise labels $\epsilon$. These pairs are used to train a score estimator~\cite{song2020score} $\epsilon_{\theta}$ usually based on the UNet architecture. The score estimator can serve as an effective approximation of the score function $\nabla \log p(x)$ which directs the inverse denoising process to generate new data samples. 

With distinguished capabilities in synthesizing images, the Latent Diffusion Model (LDM)~\cite{rombach2022high} is selected as the backbone of our method. The LDM employs a pre-trained AutoEncoder to transform the diffusion process from pixel space to latent space and integrates a conditional branch, facilitating faster training and more flexible embedding of conditions. Specifically, the pre-trained Encoder $\mathcal{E}(\cdot)$first encodes images into latent space $z = \mathcal{E}(x)$. Following this, the score estimator network $\epsilon_\theta$ is trained by taking the latent $z$, step $t$ and conditions $c$ as input to predict the noise labels:

\begin{equation}
    \min _\theta \mathbb{E}_{z ={\mathcal{E}(x)}, \epsilon \sim \mathcal{N}(0, I), t \sim \mathrm{U}(1, T)}\left\|\epsilon-\epsilon_\theta\left(z_t, t, c\right)\right\|_2^2
\end{equation}
For text to image generation task, condition $c$ is usually the text embedding generated from text prompt $y$ through a tokenizer and a pretrained CLIP~\cite{radford2021learning} model $c = \tau(y)$. The intermediate noisy latent $z_t$ is generated through the formula~\cite{ho2020denoising}:
\begin{equation}
    z_{t}=\sqrt{\bar{\alpha}(t)} z_{0}+\sqrt{1-\bar{\alpha}(t)} \epsilon, \epsilon \sim N(0, I)
\end{equation}
$\bar{\alpha}$ is the cumulative product of the noise coefficients at each step. During the sampling process, the trained score estimator takes random Gaussian noise as input, along with text embedding as condition. It progressively predicts the noise added at each step, completing the denoising process to obtain $\hat{z}_0 $. The final image is obtained by the pretrained decoder $\hat{x}_0 = \mathcal{D}({\hat{z}_0})$.

\subsubsubsection{\textbf{Textual Inversion.}} Textual inversion~\cite{gal2022image} introduces a new paradigm to T2I generation models, allowing the model to learn a new concept by setting a placeholder token "[C]" and obtaining the corresponding text embedding $\hat{v}$ as a learnable vector. This vector is then trained and optimized using a few images represent this new concept:
\begin{equation}
    \hat{v}=\underset{v}{\arg \min } \mathbb{E}_{z=\mathcal{E}(x), \epsilon \sim \mathcal{N}(0, I), t \sim U(1, T)}\left\|\epsilon-\epsilon_\theta(z_t, t, v)\right\|_2^2
\end{equation}
During training, the network parameters are all fixed, only the embedding is optimized.

\subsubsubsection{\textbf{DDIM Sampling and Inversion.}} Inversion is an effective method for finding the corresponding noise map of an image and achieving training-free control during the generation process. DDIM inversion is widely used due to its clear principles and easy implementation. The DDIM sampling process is~\cite{song2020denoising}:
\begin{equation}
    z_{t-1}=\sqrt{\bar{\alpha}_{t-1}} \frac{z_t-\sqrt{1-\bar{\alpha}_t} \epsilon_\theta\left(z_t, t, c\right)}{\sqrt{\bar{\alpha}_t}}+\sqrt{1-\bar{\alpha}_{t-1}} \epsilon_\theta\left(z_t, t, c\right)    
\end{equation}
By simply assuming $z_{t-1}\approx z_t$ and rewriting the sampling process in reverse direction, the following DDIM Inversion~\cite{song2020denoising} formula is given:
\begin{equation}
    {z}_t=\sqrt{\bar{\alpha}_t} \frac{{z}_{t-1}-\sqrt{1-\bar{\alpha}_{t-1}} \epsilon_\theta\left({z}_{t-1}, t, c\right)}{\sqrt{\bar{\alpha}_{t-1}}}+\sqrt{1-\bar{\alpha}_t} \epsilon_\theta\left({z}_{t-1}, t, c\right) 
\end{equation}
Unlike direct noise addition, the DDIM Inversion allows for the original information of the image to be well preserved, enhancing the stability in the subsequent generation process.

\subsection{Overall Pipeline}
Given a computer-rendered fashion image $x_{cg}$, the goal of our method is to transform it into a corresponding realistic image $x_r$ while preserving the garment's detailed textures. Defining realism and helping model understand what is "realistic" remains an open question. The challenge can be divided into two sub-tasks: one is making the fashion image appear realistic by enhancing aspects like wrinkles, lighting and color, which reflect true-to-life expressions. Another one is to maintain the texture details of the garment to achieve fine-grained, controllable generation.

As shown in Fig. \ref{method}, our method comprises two stages: Domain Knowledge Injection (DKI) and Realistic Image Generation (RIG). During the DKI phase, we infuse the model with information from both the rendered and realistic domains through fine-tuning and domain inversion. In the subsequent generation phase, we utilize negative domain embedding $v_{nd}$ to stimulate the model's potential for generating realistic images and employ self-attention control to preserve texture details. For a better understanding, details will be further elaborated in Section \ref{sec:DKI} and Section \ref{sec:RIG}.
\subsection{Domain Knowledge Injection}
\label{sec:DKI}
\subsubsubsection{\textbf{Target Domain Knowledge Injection}}
To enhance the ability of the base SD model $\epsilon_\theta$ to generate realistic images, especially concerning the appearance of garments and models, we use real studio-shot images $x_{tr}$ to fine-tune the base model. This process injects real domain information into the model, thereby increasing its potential to generate authentic visual details, the fine-tuning process can be formulated as:

\begin{equation}
    {\epsilon_\theta^*}=\underset{\epsilon_\theta}{\arg \min } \mathbb{E}_{z=\mathcal{E}(x_{tr}), \epsilon \sim \mathcal{N}(0, I), t \sim U(1, T)}\left\|\epsilon-\epsilon_\theta(z_t, t, v_{tr})\right\|_2^2
\end{equation}
where $\epsilon_\theta$ is the pretrained SD model, $v_{tr}$ is the embedding of the text description of the $x_{tr}$.

\subsubsubsection{\textbf{Source Domain Knowledge Injection}} For the source domain rendered data, we hope that the model can understand its characteristics and deviated from the rendered data manifold as much as possible during the generation process. After the first step fine-tuning, we assume that the model has already enhanced its representation of the real domain manifold. If we can make the model deviate from the  rendered data manifold, it can better express the characteristics of realistic images.
\begin{wrapfigure}[15]{r}{0.6\textwidth}
    \begin{center}
    \includegraphics[width=0.6\textwidth]{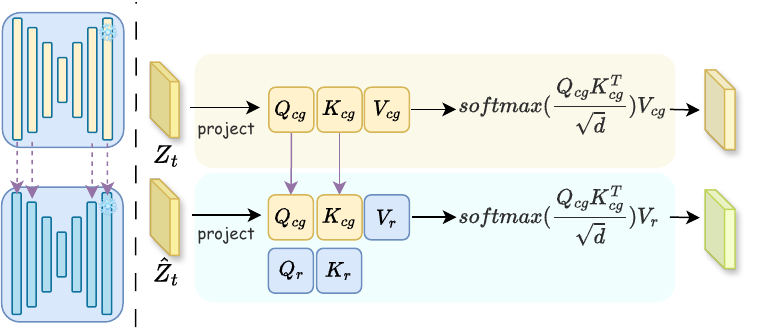} 
    \end{center}
    \caption{The diagram of Texture-preserving Attention Control (TAC).}
    \label{attn}           
\end{wrapfigure}
Inspired by the concepts of Textual Inversion and Classifier-Free Guidance (CFG) with negative prompts, we expand the concept of Textual Inversion to Domain Inversion. We train a negative domain embedding on a fine-tuned base model using a large number of rendered images. This negative domain embedding guides the model to avoid certain content, here is the rendered domain characteristics, during the generation process.

Specifically, given that textual descriptions of what is real and rendered are limited, it is difficult to guide the model to generate images with satisfactory realism or to precisely direct it not to produce images with a rendered feel using text prompts only. Therefore, we consider using negative domain embedding $v_{nd}$ trained on a large number of rendered images for guidance to inject the rendered domain knowledge to the model. It's worth nothing that unlike textual inversion, which typically optimizes a small embedding space with few images to represent a specific concept, such as a particular object in personalized generation or an easily expressible style. The concept of rendered domain in our task is much more general. Using a small embedding space corresponding to few images to represent this would easily lead to over-fitting to the content of the training images. We use the largest available embedding size to train the negative domain embedding, which is corresponding to the placeholder token size of 75:
\begin{equation}
    \hat{v}_{nd}=\underset{v}{\arg \min } \mathbb{E}_{z=\mathcal{E}(x_{cg}), \epsilon \sim \mathcal{N}(0, I), t \sim U(1, T)}\left\|\epsilon-\epsilon_\theta^*(z_t, t, v)\right\|_2^2
\end{equation}
During the training of negative domain embedding, we freeze the parameters in the fine-tuned model $\epsilon_\theta^*$, and find the $v_{nd}$ through direct optimization with a certain number of rendered images. 

\subsection{Realistic Image Generation}
\label{sec:RIG}
\subsubsubsection{\textbf{Negative Embedding Guidance}}
After domain knowledge injection, we can use the negative domain embedding to guide the model in generating realistic images. During each denoising step, the negative domain embedding guidance is defined by:
\begin{equation}
\label{eq8}
    \tilde{\epsilon}_\theta^*\left(z_t, t, v_{nd}\right)=w \cdot \epsilon_\theta^*\left(z_t, t, v_\varnothing\right)+(1-w) \cdot \epsilon_\theta^*\left(z_t, t, v_{nd}\right)
\end{equation}
where $v_\varnothing$ denotes the embedding of Null text. With a guidance scale $w$ larger than 1, the negative domain embedding becomes effective. Unlike traditional CFG guidance, here we do not use any positive prompts processed through CLIP to obtain the embedding as conditions. Instead, we directly employ a Null text embedding. The initial noise latent is obtained through DDIM inversion of the given rendered image. During the denoising process, we replace the CLIP conditioning branch, since the negative domain embedding is trained on a fine-tuned model, it can interact more effectively with the base model's latent space. This consistency allows for more precise adjustments in the latent manifold compared to embedding derived from text via CLIP.

\subsubsubsection{\textbf{Texture-preserving Attention Control (TAC)}} 
Inspired by previous work~\cite{tumanyan2023plug, liu2024towards}, the attention features in the diffusion UNet, which includes both cross attention and self attention, hold rich information critical for generating the new images. Cross attention typically handles the attributes and semantics of the generated image, while self-attention maps play a crucial role in preserving geometric shapes and intricate details. The initial noise latent $\hat{Z}_t$ derived from the DDIM inversion of the original rendered image can be used in unconditional generation and extract the texture related attention features as shown in Fig. \ref {attn}. 
However, directly replacing all self-attention maps can lead to a decrease in the realism of the generated images. We argue that this is because the attention map contains both the texture details of the garment and the general rendered domain characteristics. Therefore, we propose to control the self attention feature only in the shallow layers of the denoising UNet to decouple the texture details feature from the general rendered domain features. During the implementation, we also find that in the deep feature spaces with higher downsampling rates, it becomes challenging to identify features related to the texture details. Thus, our TAC is defined as:
\begin{equation}
\label{eq9}
    \widehat{Q^t},\widehat{K^t}=TAC\left(Q_{cg}^t,K_{cg}^t, Q_r^t, K_r^t, t\right)= \begin{cases} Q_{cg}^t,K_{cg}^t & \text { if } t < \gamma T, f > F \\ Q_r^t, K_r^t & \text { otherwise }\end{cases}
\end{equation}
where $\gamma$ is the parameter that indicates how many steps before the TAC should be applied and $f$ is the feature size of different layers, only those layers exceeding the specified size $F$ undergo TAC, particularly in the shallow layers. Specifically, the cg-domain self-attention features are derived from the reverse sampling process starting from the noisy latent, which is obtained by performing DDIM inversion on the input image latent. In contrast, the r-domain self-attention features differ due to the incorporation of negative domain guidance and the self-attention injection.


\begin{figure}        
\centering
\includegraphics[width=\textwidth]{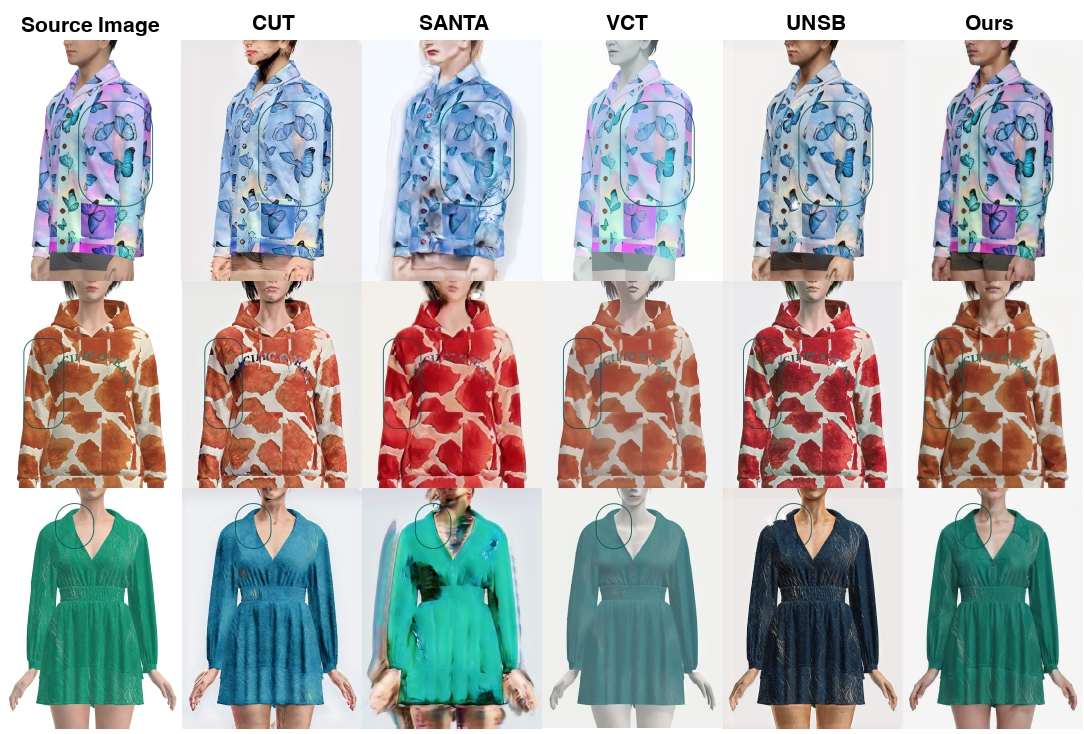} 
\caption{Results on our proposed SynFashion Dataset. (\textbf{Please zoom in for details.)}}
\label{result_garment}                     
\end{figure}

\begin{figure}        
\centering
\includegraphics[width=\textwidth]{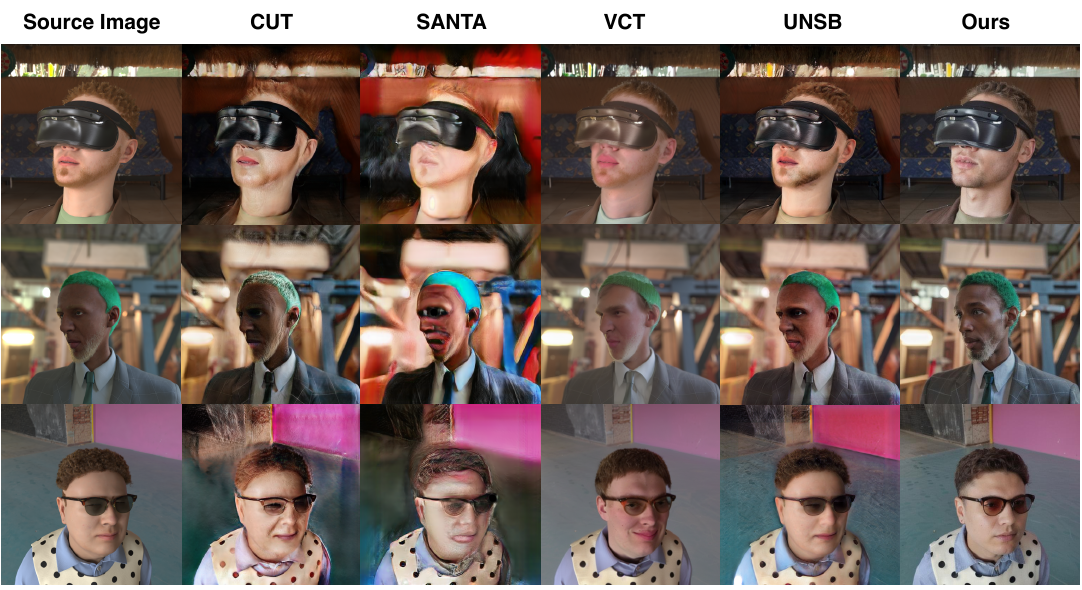} 
\caption{Results on the Face Synthetics dataset. (\textbf{Please zoom in for details.)}}
\label{result_face}                     
\end{figure}

\section{Experiments}
\subsection{Datasets}
To evaluate our method and conduct comprehensive comparisons, we introduce a high-quality rendered fashion image dataset, named Synthetic Fashion (SynFashion), with the professional garment design software Style3D Studio. SynFashion consists of 10k rendered images in 20 categories, including pants, T-shirt, lingerie and swimwear, half skirt, hoodie, coat, jacket, set, home-wear, hat, Hanfu, jeans, shorts, down jacket, vest and camisole, shirt, suit, dress, sweater and trench coat. For each category, we use Style3D Studio to build 10 to 40 projects in different 3D geometry with corresponding texture and design, and then randomly sample several new textures to change its appearance. There are overall 375 projects in 3D and 500 additional texture collected from Internet. For each textured 3D geometry, we render four views, including front, back, and two randomly sampled views. After rendering, we crop the enlarged garment area of each image and resize it to 768 $\times$ 1024. Due to legal issues, some of the images contain a digital human figure but not the complete face. To supplement the evaluation on rendered human faces, we also conduct experiments on the public available Face Synthetics dataset~\cite{wood2021fake} with its first 10k images.

\subsection{Implementation Details}
\subsubsubsection{\textbf{Implementation.}} We implement our method with pretrained Stable Diffusion (SD) model and finetune the base model with 2500 realistic images at a $1024\times1024$ resolution for source domain knowledge injection. The finetuning uses images from iMaterialist (Fashion) 2019 FGVC dataset~\cite{imaterialist-fashion-2019-FGVC6}, based on the publicly available SD v1.5, and is conducted on 2 RTX 4090 with a batch size of 6. Based on the finetuned model, we train our negative domain embedding with 2500 rendered images on a single RTX 4090 with a batch size of 1. The rendered images are resized to the resolution of $512\times512$. The placeholder embedding size is 75 and the learning rate is 5e-4. During sampling, we perform DDIM sampling with default 50 denoising steps with a denoising strength of 0.3 as default. The $\gamma$ is set to 0.9 as default, which means that the TAC is performed on the first 90\% of sampling steps. Only the attention maps in the first and second shallow layers are used for TAC. Note that the denoising strength and $\gamma$ may be changed to obtain different level of image translation. We compare our method with three state-of-the-art unpaired image-to-image translation method, CUT, SANTA and UNSB, and one diffusion-based style transfer method VCT. For CUT, SANTA and UNSB, we train the models for about 400 epochs following the official code with same training data.

\subsection{Results}
\subsubsubsection{\textbf{Qualitative Results}} 
Fig. \ref{result_garment} and Fig. \ref{result_face} show the visual comparison between our method, CUT~\cite{park2020contrastive}, SANTA~\cite{xie2023unpaired}, UNSB~\cite{kim2023unpaired} and VCT~\cite{cheng2023general} on the SynFashion and Face Synthetics datasets. As can be seen from the figures, both the CUT and SANTA methods exhibit some degree of image degradation and fail to effectively learn the concept of image realism from data across rendered and real domains, thus enable to generate realistic images. The diffusion based style transfer method VCT maintains image quality but fails to extract realistic image features from the guidance image, also resulting in the loss of image details. Compared to previous methods, the UNSB method achieves better consistency in terms of content, but like CUT and SANTA, it performs poorly in maintaining color fidelity and the realism effect is not good. The proposed method effectively enhances the overall realism of the image, particularly in capturing the facial and hand features of models, as well as the texture and wrinkle details of the garment.
\subsubsubsection{\textbf{Quantitative Results}}
The absence of ground truth for rendered-to-real translation and domain gap between the source rendered and target real domains make quantitative evaluation challenging. 
\begin{table}
\centering
\caption{Quantitative comparisons on Face Synthetics and SynFashion datasets.}
\scalebox{0.8}{
\begin{tabular}{@{}lccccccccc@{}}
\toprule
\textbf{Dataset} & \multicolumn{3}{c}{\textbf{Face Synthetics}} & \multicolumn{3}{c}{\textbf{SynFashion}} \\ \cmidrule(lr){2-4} \cmidrule(lr){5-7}
                                   & KID$\downarrow$(std)  & LPIPS$\downarrow$(std)  & SSIM$\uparrow$(std)    & KID$\downarrow$(std)   & LPIPS$\downarrow$(std)  & SSIM$\uparrow$(std) \\ \midrule
CUT~\cite{park2020contrastive}   & 80.553 (2.447)    & 0.365 (0.073)   & 0.664 (0.079)   & \underline{59.238} (1.599)   & 0.170 (0.060) & 0.847 (0.067)\\
SANTA~\cite{xie2023unpaired}     & 90.390 (2.929)      & 0.387 (0.079)  & 0.618 (0.104)     & 61.636 (1.628)  & 0.294 (0.067)  & 0.741 (0.082)\\
VCT~\cite{cheng2023general}      & \underline{74.445} (2.273)   & \textbf{0.096} (0.027)   & 0.807 (0.072)   & 59.489 (1.499) & 0.178 (0.058) &0.807 (0.085)  \\
UNSB~\cite{kim2023unpaired}     & 76.389 (2.465)   & 0.229 (0.069)    & \underline{0.818} (0.070)  & 59.496 (1.453) & \underline{0.130} (0.040) & \textbf{0.891} (0.054) \\
Ours                            & \textbf{73.871} (1.973)    & \underline{0.121} (0.035)    & \textbf{0.831} (0.068)  &\textbf{54.720} (1.362) & \textbf{0.067} (0.025) & \underline{0.881} (0.055)\\
\bottomrule
\end{tabular}}
\label{table-1}
\end{table}

Following the previous work~\cite{richter2022enhancing}, we use KID to evaluate the realism of the generated images and the average SSIM and LPIPS to assess content similarity. For each dataset, we use the 7500 testing result images from each method and calculate the KID against the realistic images and the SSIM/LPIPS against the rendered images. As shown in Tab. \ref{table-1}, our method shows significant improvements in terms of realism as well as overall texture and content consistency. The standard deviations here show the variance over test inputs for a fixed model to demonstrate the stability and generalization ability.

\begin{table}
    \centering
    \caption{User studies on overall realism, image quality and consistency. The table shows the percentage of votes that existing methods are preferred to ours.}
    \scalebox{0.8}{
    \begin{tabular}{@{}lccccccccc@{}}
    \toprule
    \textbf{Dataset} & \multicolumn{3}{c}{\textbf{Face Synthetics}} & \multicolumn{3}{c}{\textbf{SynFashion}} \\ \cmidrule(lr){2-4} \cmidrule(lr){5-7}
                                    & Overall Realism   & Image Quality  & Consistency   & Overall Realism   & Image Quality   & Consistency  \\ \midrule
    CUT    & 0.529\%        & 0.529\%    & 13.175\%       & 8.994\%           & 6.878\%  & 16.931\% \\
    SANTA  & 0.922\%        & 1.383\%    & 12.304\%       & 3.333\%           & 5.238\%  & 11.571\% \\
    VCT    & 5.952\%        & 14.286\%   & 20.714\%       & 2.041\%           & 6.122\%  & 18.367\% \\
    UNSB   & 4.511\%        & 6.767\%    & 21.278\%       & 9.821\%           & 9.821\%  & 26.607\% \\
    \bottomrule
    \end{tabular}}
\label{tab-user_study}
\end{table}

\subsubsubsection{\textbf{User Studies}}
We adopt user studies to provide more quantitative insight into perceived realism, image quality, and consistency to input rendered images. We follow StyleDiffusion~\cite{wang2023stylediffusion} in style-transfer and compare our method to previous works in pairs. Specifically, we randomly sample 100 image pairs from each dataset for user evaluation. Each pair contains one image generated by our method and a corresponding image generated by another comparison method, presented side by side in random order. Users are asked to assess the images based on three criteria: 1) which result appears more realistic, 2) which result demonstrates overall better image quality, and 3) which result shows better consistency with the reference image.

We collected approximately 2,000 votes per question from 20 users and present the percentage of votes where existing methods were preferred over ours in the Tab.~\ref{tab-user_study}. Lower percentages indicate that our method was favored over the competitors. Our approach garnered a strong preference in terms of overall realism and image quality, while also showing a clear advantage in maintaining consistency with the reference images.

\begin{figure}[t!]
\centering
\includegraphics[width=\textwidth]{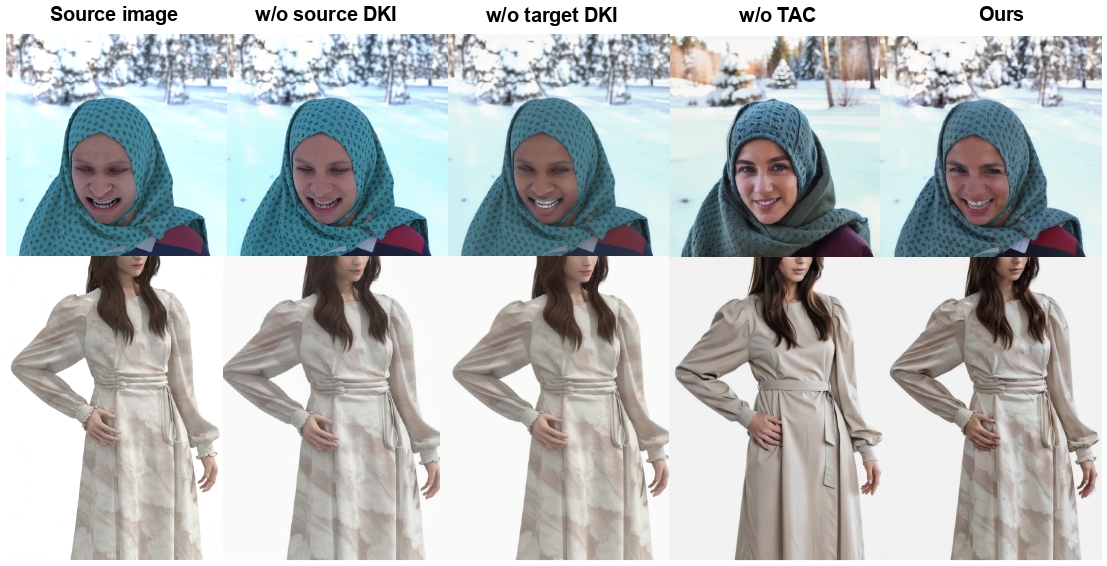}
\caption{Visual examples of ablation study in a drop-one-out manner. (DKI: Domain Knowledge Injection. TAC: Texture-preserving Attention Control.)}
\label{ablation_modules}                     
\end{figure}

\begin{table}[t!]
    \centering
    \caption{Ablation study in a drop-on-out manner.}
    \scalebox{0.8}{
    \begin{tabular}{@{}lccccccccc@{}}
    \toprule
    \textbf{Dataset} & \multicolumn{3}{c}{\textbf{Face Synthetics}} & \multicolumn{3}{c}{\textbf{SynFashion}} \\ \cmidrule(lr){2-4} \cmidrule(lr){5-7}
                                 & KID$\downarrow$(std)  & LPIPS$\downarrow$(std) & SSIM$\uparrow$(std)   & KID$\downarrow$(std)   & LPIPS$\downarrow$(std)  & SSIM$\uparrow$(std) \\ \midrule
    w/o source DKI   & 77.376 (2.063)    & 0.107 (0.029)   & 0.857 (0.059)            & 58.520 (1.902)  & 0.059 (0.019)  & 0.903 (0.065) \\
    w/o target DKI   & 78.927 (2.134)    & 0.114 (0.031)   & 0.845 (0.063)            & 60.186 (1.623)  & 0.064 (0.022)  & 0.897 (0.056) \\
    w/o TAC          & 69.349 (1.485)    & 0.253 (0.070)   & 0.720 (0.085)            & 51.392 (1.083)  & 0.183 (0.047)  & 0.794 (0.074) \\
    Ours             & 73.831 (1.973)    & 0.121 (0.035)   & 0.831 (0.068)            & 54.720 (1.362)  & 0.067 (0.025)  & 0.881 (0.055) \\
    \bottomrule
    \end{tabular}}
\label{tab-ablation}
\end{table}

\begin{figure}[t!]
\centering
\includegraphics[width=0.79\textwidth]{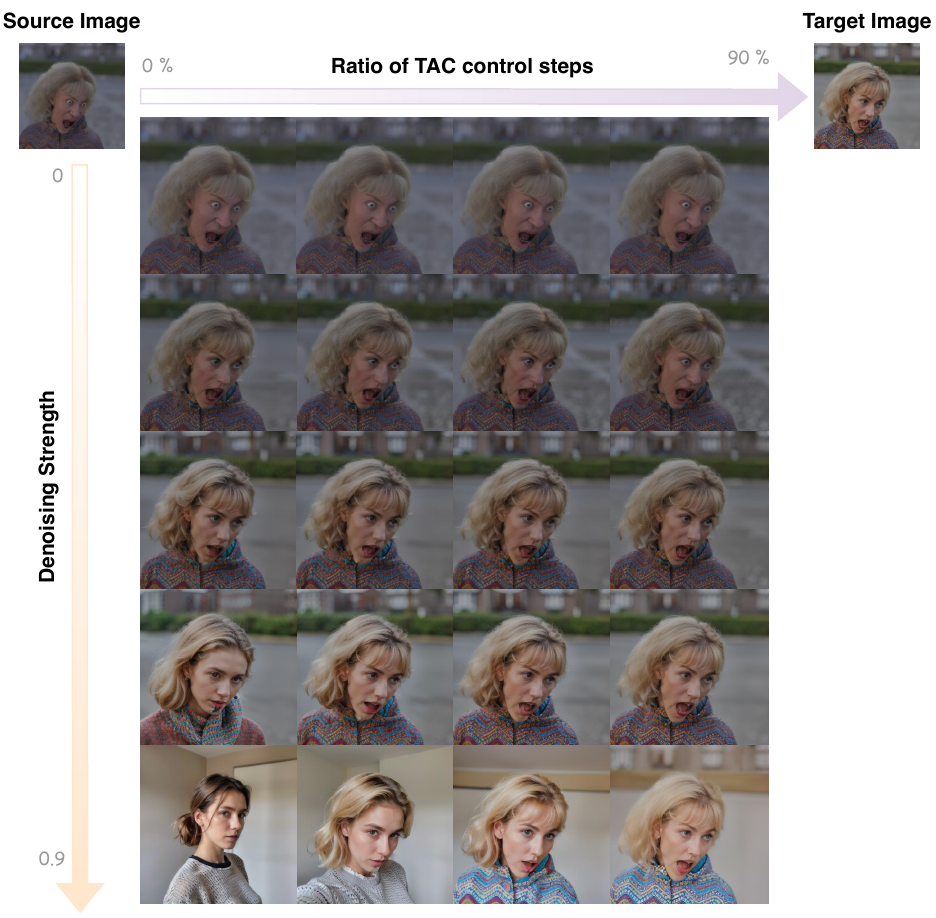}
\caption{A visual example of tuning TAC ratio and denoising strength.}
\label{ablation_tac}                     
\end{figure}

\subsection{Ablation Study and Further Analysis}
We conduct ablation study on two datasets in a drop-one-out manner and evaluate the performance of each module in the proposed method during inference and analyze the impact on the final results. 
As shown in Fig. \ref{ablation_modules}, without source DKI (embedding), the fine-tuned base model tends to recover the input rendering image with DDIM inversion. Without target
DKI  (fine-tuning), the rendering effect slightly decreases but the output is still not real enough due to lack 
of concentrated knowledge on real human and clothing. Without TAC, the semantic structure such as face identity and clothing design can significantly deviate from the input. The quantitative results are in Tab.~\ref{tab-ablation}.

Fig. \ref{ablation_tac} shows the trade-off between image realism and texture preservation. With a high denoising strength, the generated images resemble realistic images more closely but retain fewer details from the original rendered image. Increasing the TAC ratio helps to better preserve the texture details and facial features. Unlike other content preservation techniques such as inpainting, which can lead to potential visual incoherence, our TAC seems to blend the attention features smoothly into the generation process and cause no obvious coherence issues.

\section{Conclusion}
In this paper, we introduce a novel diffusion-based framework for rendered-to-real fashion image translation and create a high-quality rendered fashion image dataset (SynFashion), which includes 10k images with multiple classes. With Domain Knowledge Injection (DKI) and Texture-preserving Attention Control (TAC), our method can successfully translate the rendered fashion image into its realistic counterpart with significant realism improvement and texture details preservation. Extensive experimental results demonstrate the superiority and effectiveness of our method.

\begin{table}[t!]
    \centering
    \caption{Comparison of memory required and testing time across different methods.}
    \scalebox{0.8}{
    \begin{tabular}{lccccc}
        \toprule
        & CUT & SANTA & VCT & UNSB & Ours \\
        \midrule
        Memory Required (GB) & 3.3 & 4.5 & 22 & 7.4 & 7.7 \\
        Testing Time (s) & 0.38 & 0.33 & 62.47 & 0.53 & 7.98 \\
        \bottomrule
    \end{tabular}}
\label{tab-time}
\vspace{-0.3cm}
\end{table}

\subsubsubsection{\textbf{Limitations and social impacts}}
While our method achieves superior results on this challenging task, there are still several problems to be further explored. In this work, we simply use DDIM inversion to extract texture-related attention features. However, the inversion process slows down the generation, requiring approximately one minute to translate an image with a resolution of $768 \times 1024$. This could potentially be accelerated by recent inversion-free methods. We test the inference time and resource consumption for a 512x512 image on an RTX 3090, as shown in Tab.~\ref{tab-time}. Note that comparing to VCT, which is also based on diffusion, our method takes much less memory and time during testing as we do not need to perform additional optimization for each testing image. Our method cannot handle real-time applications for now, but has potential for improvement with future integration with SD Turbo or SD Lightning. Additionally, for different images, finding the optimal balance between the TAC ratio and denoising strength may require more empirical refinements to achieve the best result. Due to limitations on computational resources, experiments were not conducted on more advanced models such as SDXL~\cite{podell2023sdxl}. Given that our method is based on SD1.5 and for human-related content generation, potential negative societal impacts of exploiting this method could be violation of portrait rights, racial bias, or inappropriate content in generation when the denoising strength is high. Relative solutions can include but are not limited to using authorized, diverse and balanced training data and training detection models to prevent inappropriate content generation.

\section{Acknowledgments}
This work is supported in part by the National Key Research and Development Program of China (No: 2021YFF0501503) and by the Talent Program of Zhejiang Province (No: 2021R51004).

\newpage
\bibliographystyle{unsrt}
\bibliography{neurips_2024.bib}

\clearpage
\appendix

\section{Appendix / supplemental material}

\begin{figure}[H]
\centering
\includegraphics[width=\textwidth]{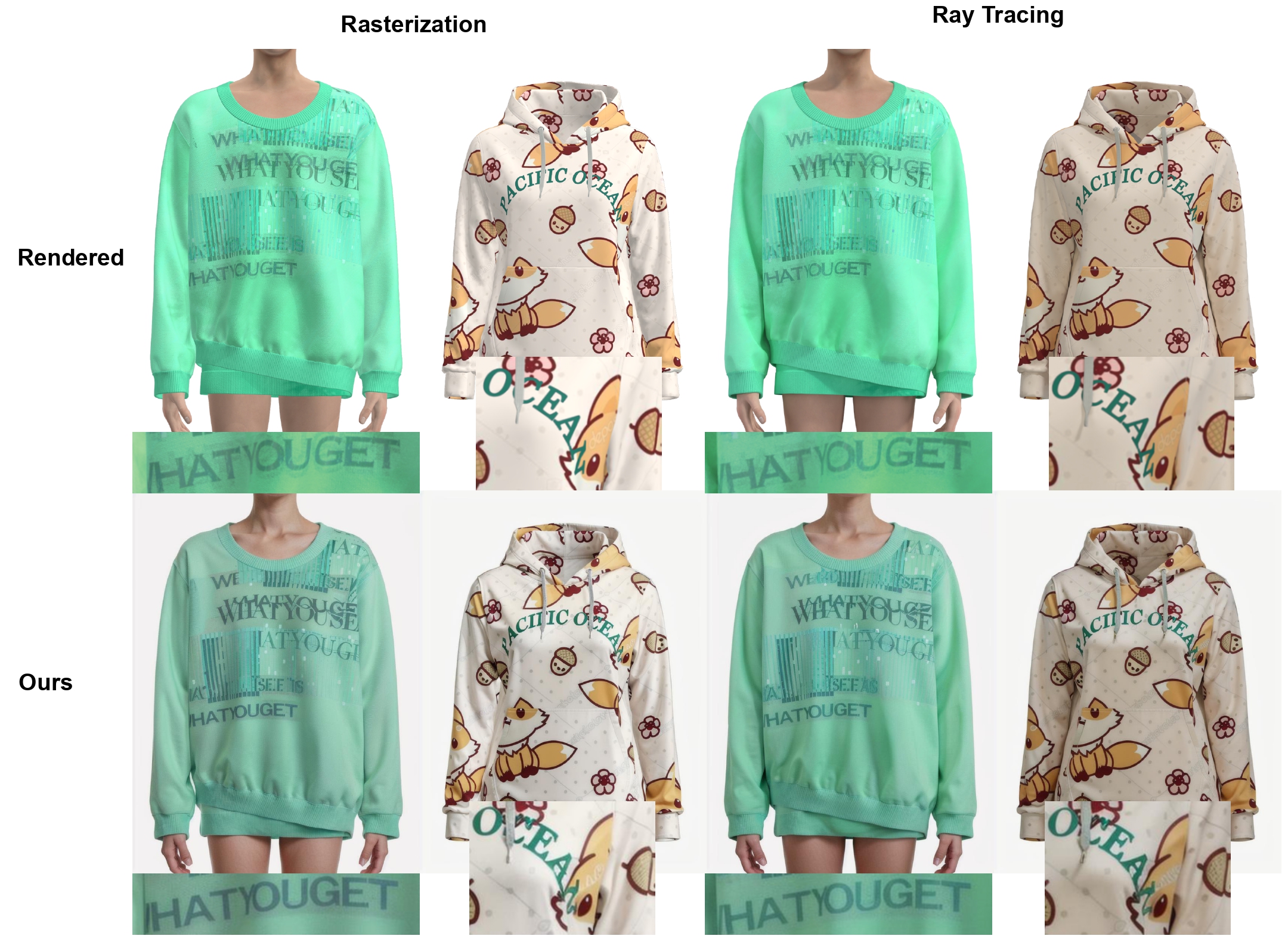} 
\caption{Results on textual textures and different rendering inputs.}
\label{Rendering}                     
\end{figure}

\subsection{More implementation details}

More details of our RIG algorithm is shown in Algorithm.~\ref{Algorithm1}.

\begin{algorithm}[H]
    \caption{Realistic Image Generation}
    \begin{algorithmic}[1]
    \State \textbf{Inputs:}
    \State $x_{cg} \gets$ source rendered image
    \State $v_{nd} \gets$ negative domain embedding
    \State $\gamma, F \gets$ step threshold, feature size threshold
    \State \textbf{Algorithm:}
    \State $z_0 = \mathcal{E}({x_{cg}})$
    \State $\hat{z}_T \gets \text{DDIM-inv}(z_0)$
    \State $z_T \gets \hat{z}_T \quad \text{// starting from same seed}$
    \For{$t = T$ to $1$}
        \State $z_{t-1}, Q_{cg}^t, K_{cg}^t \gets \text{DDIM-samp}(z_t)$
    
        \If{$t < \gamma \And f > F$}
            \State $\hat{z}_{t-1} \gets \tilde{\epsilon}_\theta ^*(\hat{z}_t, t, v_{nd})\{Q_r^t \gets Q_{cg}^t; K_r^t \gets K_{cg}^t\}$
        \Else
            \State $\hat{z}_{t-1} \gets \tilde{\epsilon}_\theta ^*(\hat{z}_t, t, v_{nd})$
        \EndIf
    \EndFor
    \State \textbf{Output:} $x_{r} \gets \mathcal{D}(\hat{z}_0)$
    \end{algorithmic}
\label{Algorithm1}
\end{algorithm}

\subsection{Results on textual textures and different rendering inputs}
As for rendering baselines, we build the 3D projects with Style3D Studio and use its integrated rendering tool based on rasterization. Using UE5 could potentially improve the rendering quality but will not diminish the effectiveness of our method. To verify this, we use more advanced rendering techniques via ray tracing (based on V-ray) to obtain rendered images, and our method consistently demonstrates its advantages in realism. Two visual examples are shown in Fig.~\ref{Rendering}.

\subsection{More results of realistic image translation}
\begin{figure}        
\centering
\includegraphics[width=\textwidth]{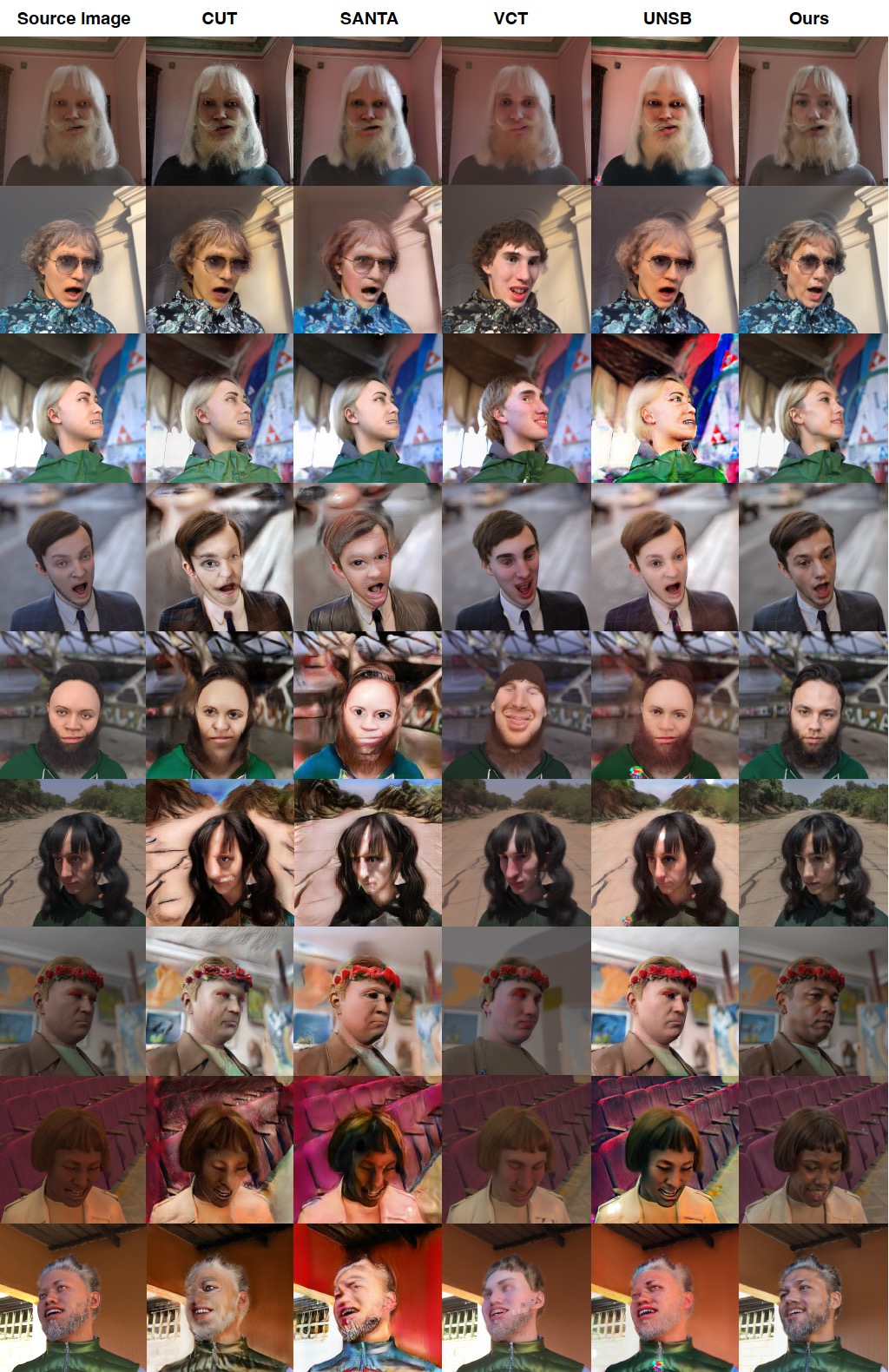} 
\caption{More comparison results on Face Synthetics dataset.}
\label{More_faces}                     
\end{figure}

\begin{figure}        
\centering
\includegraphics[width=\textwidth]{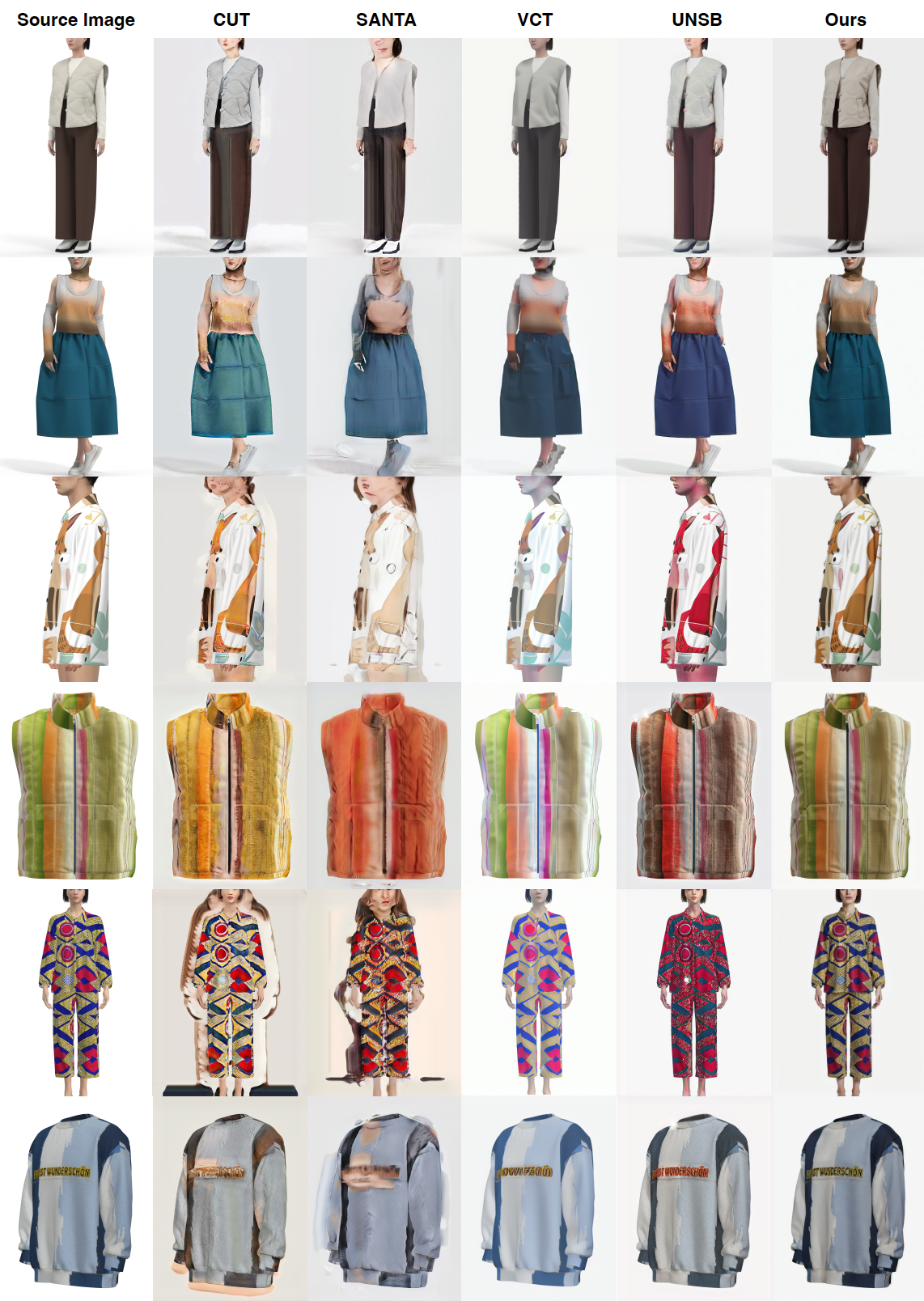} 
\caption{More comparison results on SynFashion dataset.}
\label{More_garments}                     
\end{figure}

To further verify the performance of the proposed method in realistic translation tasks, additional experiments were conducted using the collected SynFashion dataset and the Face Synthetics dataset. The results are illustrated in Figure .\ref{More_faces} for Face Synthetics and Figure .\ref{More_garments} for SynFashion.

\subsection{More details of collected SynFashion dataset}
\begin{figure}        
\centering
\includegraphics[width=\textwidth]{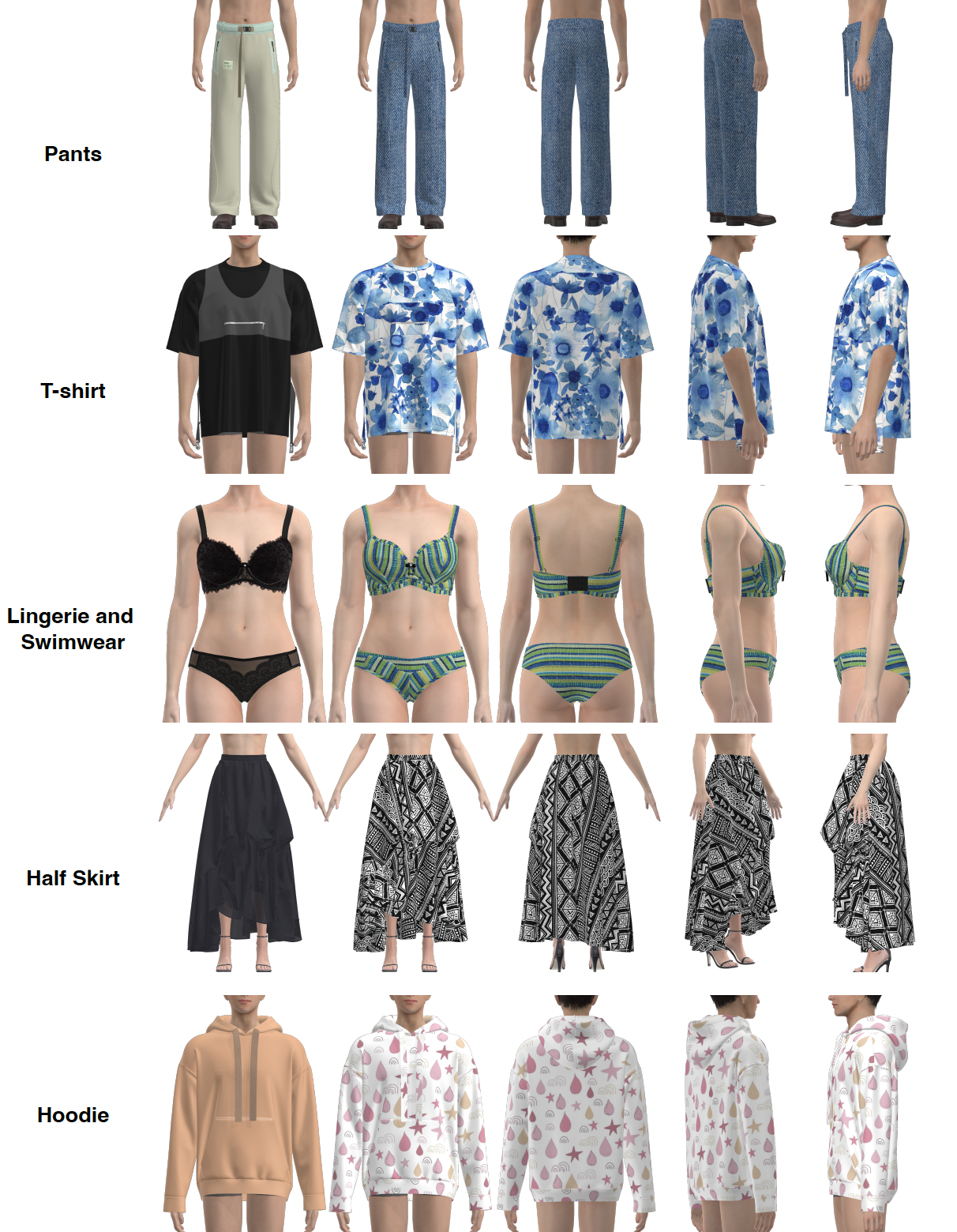} 
\caption{Examples of collected SynFashion dataset (Part 1).}
\label{dataset_1}                     
\end{figure}

\begin{figure}        
\centering
\includegraphics[width=\textwidth]{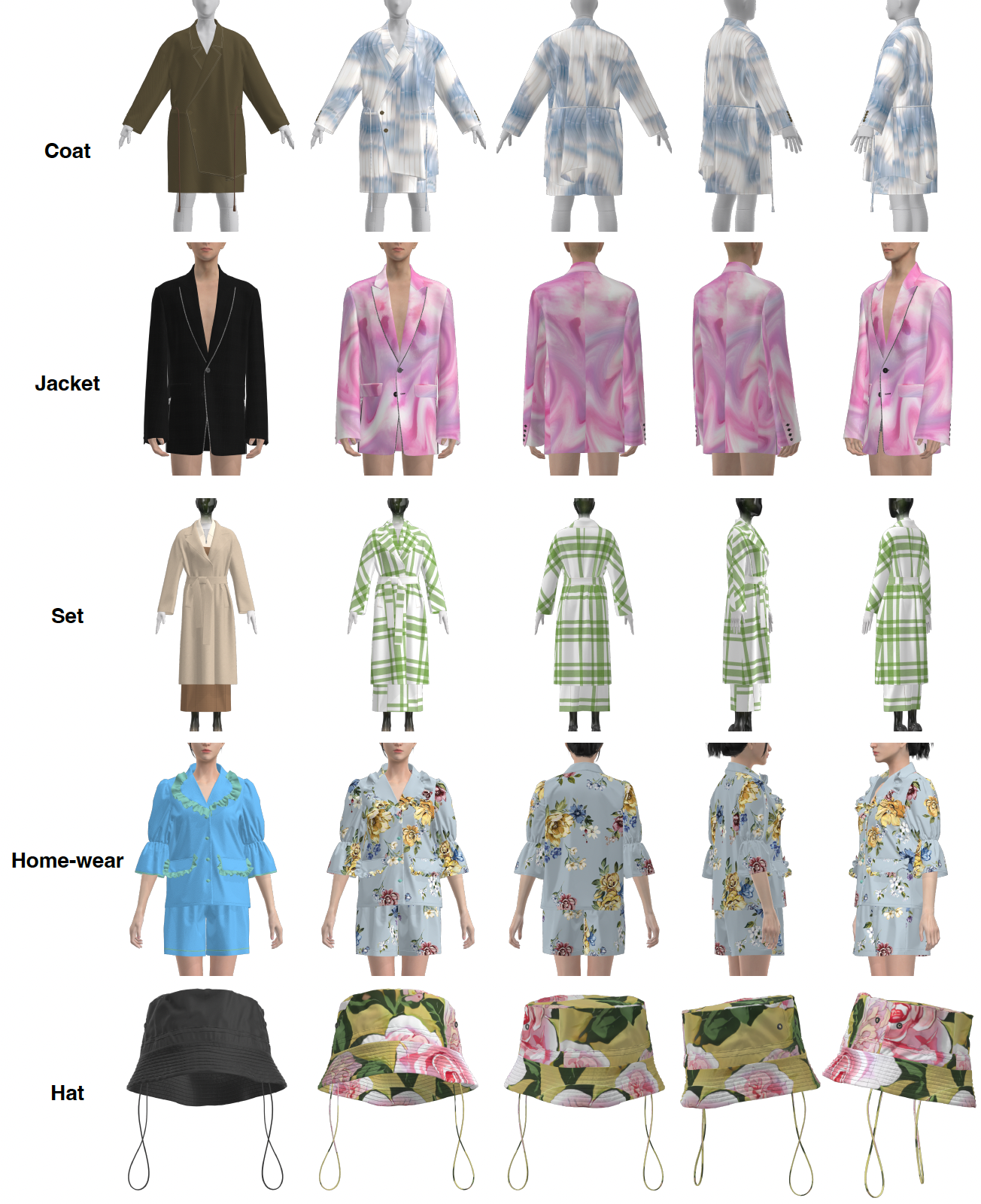} 
\caption{Examples of collected SynFashion dataset (Part 2).}
\label{dataset_2}                     
\end{figure}

\begin{figure}        
\centering
\includegraphics[width=\textwidth]{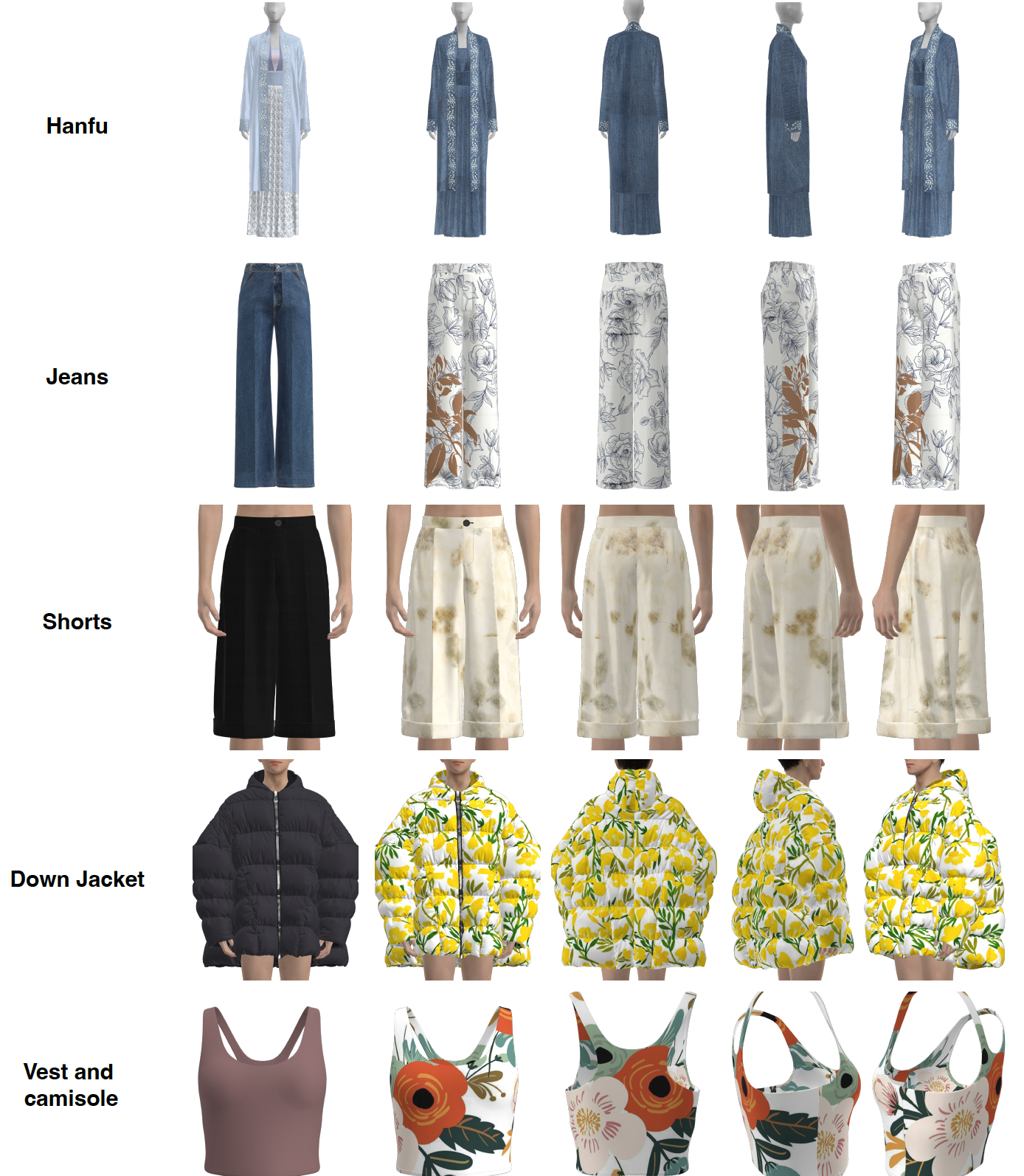} 
\caption{Examples of collected SynFashion dataset (Part 3).}
\label{dataset_3}                     
\end{figure}

\begin{figure}        
\centering
\includegraphics[width=\textwidth]{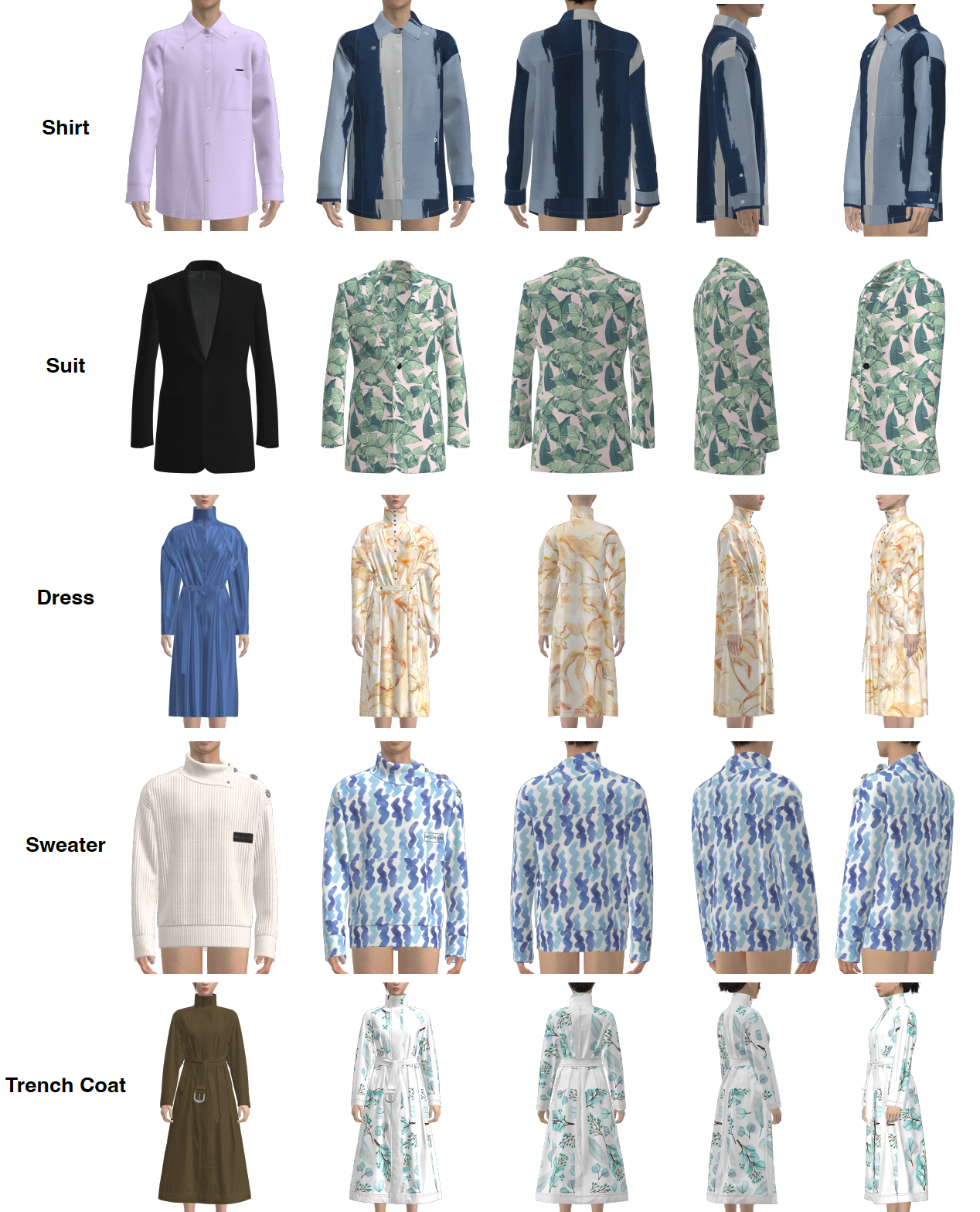} 
\caption{Examples of collected SynFashion dataset (Part 4).}
\label{dataset_4}                     
\end{figure}

Figure .\ref{dataset_1}, Figure .\ref{dataset_2}, Figure .\ref{dataset_3}, and Figure .\ref{dataset_4} provide detailed visualizations of the SynFashion dataset. The first column in each figure presents the front view of a designed 3D garment object. Various texture patterns are assigned to each garment object, and the subsequent columns show the images with four different views. The number of images in each category is shown in Table.~\ref{tab5}.

\newcolumntype{Y}{>{\centering\arraybackslash}X} 

\begin{table}[t!]
\centering
\caption{Number of images in different categories of SynFashion.}
\label{tab:numbers_part1}
\small 
\begin{tabularx}{\textwidth}{@{}Y*{8}{Y}@{}}
\toprule
Category & Pants & T-shirt & Lingerie \& Swimwear & Half Skirt & Hoodie & Coat & Jacket\\ 
\midrule
Numbers & 864 & 416 & 440 & 392 & 448 & 604 & 812\\
\midrule
Category & Set & Home-wear & Hat & Hanfu & Jeans & Shorts & Down Jacket\\ 
\midrule
Numbers & 420 & 336 & 308 & 472 & 180 & 420 & 508\\
\midrule
Category & Vest \& Camisole & Shirt & Suit & Dress & Sweater & Trench Coat\\ 
\midrule
Numbers & 388 & 476 & 672 & 416 & 1056 & 416\\
\bottomrule
\end{tabularx}
\label{tab5}
\end{table}

\end{document}